\documentclass[10pt,twocolumn,letterpaper]{article}

\usepackage{cvpr}              

\usepackage{graphicx}
\usepackage{amsmath}
\usepackage{amssymb}
\usepackage{booktabs}
\usepackage{chngpage}
\usepackage{tabularx}
\usepackage{color}
\usepackage{soul}

\usepackage{multirow}
\usepackage{float} 
\usepackage{tabularx}

\usepackage[T1]{fontenc}
\usepackage[utf8]{inputenc}
\usepackage{graphicx}
\usepackage[pagebackref,breaklinks,colorlinks]{hyperref}
\usepackage[capitalize]{cleveref}
\usepackage{breakurl}

\usepackage[pagebackref,breaklinks,colorlinks]{hyperref}

\usepackage[capitalize]{cleveref}
\crefname{section}{Sec.}{Secs.}
\Crefname{section}{Section}{Sections}
\Crefname{table}{Table}{Tables}
\crefname{table}{Tab.}{Tabs.}

\begin{document}

\title{Large Language Models in Politics and Democracy: A Comprehensive Survey}

\author{Goshi Aoki\\
    Zhejiang University, China\\
    Computer Science and Technology\\
    {\tt\small 2222135@zju.edu.cn}
}
\maketitle

\begin{abstract}
The advancement of generative AI, particularly large language models (LLMs), has a significant impact on politics and democracy, offering potential across various domains, including policymaking, political communication, analysis, and governance. This paper surveys the recent and potential applications of LLMs in politics, examining both their promises and the associated challenges. This paper examines the ways in which LLMs are being employed in legislative processes, political communication, and political analysis. Moreover, we investigate the potential of LLMs in diplomatic and national security contexts, economic and social modeling, and legal applications. While LLMs offer opportunities to enhance efficiency, inclusivity, and decision-making in political processes, they also present challenges related to bias, transparency, and accountability. The paper underscores the necessity for responsible development, ethical considerations, and governance frameworks to ensure that the integration of LLMs into politics aligns with democratic values and promotes a more just and equitable society.
\end{abstract}

\section{Introduction}
The advent of artificial intelligence (AI) has had profound implications across various domains, and politics and democracy are no exception. AI technologies have the potential to revolutionize the way political processes are conducted, from campaigning and public opinion analysis to policy-making and governance \cite{helbing2019will, helbing2019societal, Christopher2314021121}. In recent years, the emergence of large language models (LLMs) has opened up new possibilities for AI applications in politics, attracting significant attention from researchers, policymakers, and the general public \cite{bommasani2021opportunities}.

LLMs are a type of AI model that has gained significant attention for its ability to process and generate human-like text. These models, such as GPT-4 \cite{openai2024gpt4technicalreport} and Gemini \cite{geminiteam2024geminifamilyhighlycapable}, have demonstrated impressive capabilities in natural language understanding and generation tasks. LLMs are based on deep learning architectures, typically transformer models \cite{vaswani2023attentionneed}, and are trained in an unsupervised manner on large corpora of text data. Once trained, LLMs can be fine-tuned for various downstream tasks or used in a few-shot learning setting, where they can perform tasks with minimal task-specific training data \cite{brown2020languagemodelsfewshotlearners}.

The use of LLMs in politics holds immense potential to enhance efficiency, improve decision-making, and promote citizen engagement \cite{ziems2024largelanguagemodelstransform, durmus2024measuringrepresentationsubjectiveglobal}. For instance, LLMs can automate tasks such as analyzing policy documents and generating reports, freeing up time for policymakers and their staff to focus on more strategic and complex issues \cite{gunes2023multiclassclassificationpolicydocuments}. Additionally, LLMs can facilitate citizen participation by providing interactive platforms for deliberation and feedback on policy proposals \cite{kuo2024policycraftsupportingcollaborativeparticipatory, ashkinaze2024pluralsguidingllmssimulated}. However, the integration of LLMs in politics also presents challenges, particularly regarding bias, transparency, and accountability \cite{qi2024representationbiaspoliticalsample}.  It is crucial to address these challenges to ensure that LLMs are deployed responsibly and ethically in political contexts \cite{ziegler2024ai}.

This paper aims to provide a comprehensive overview of the current and potential applications of LLMs in politics, examining both the promises and challenges associated with their integration.  We will explore how LLMs are being used in various political processes, including policymaking, communication, analysis, and decision-making, drawing on recent research and studies \cite{gao2023implications, Michael20531680241236239, zhang2024electionsimmassivepopulationelection, lin2024battleagentmultimodaldynamicemulation}.   Additionally, we will discuss the ethical and societal implications of using LLMs in politics, highlighting the need for responsible development and governance frameworks \cite{caballero2024largelanguagemodelsnational, park2023aideceptionsurveyexamples}. By examining the multifaceted role of LLMs in the political field, this review seeks to contribute to a deeper understanding of how AI is shaping the future of governance and democratic participation \cite{helbing2019societal}.

\section{Understanding Large Language Models}
This section provides an overview of LLMs' technical foundations, capabilities, and key characteristics.

LLMs are trained on a huge amount of text data, including books, articles, code, and social media conversations, enabling them to develop a broad understanding of human language and generate code and text.

Some of the most popular LLMs include:

\textbf{GPT-4 (OpenAI):} A multimodal model that can input image and text and produce text outputs and images.  It excels in tasks requiring reasoning, creativity, and nuanced understanding of language \cite{openai2024gpt4technicalreport}. 

\textbf{Gemini (Google):} A multimodal model that is natively conditional and can handle many tasks with minimal modifications.  It is designed to be efficient and scalable, making it suitable for various applications with long context \cite{geminiteam2024geminifamilyhighlycapable}. 

\textbf{LLaMA (Meta):} An one of the open source LLM ranging from 7B to 65B parameters.  They are designed to democratize access to large language models and promote transparency in AI research \cite{touvron2023llama}.

LLMs are neural networks based on the transformer architecture \cite{vaswani2023attentionneed}, trained on massive text corpora through self-supervised learning. These models process text as sequences of tokens and employ attention mechanisms to capture complex relationships between words and concepts. Modern LLMs like GPT-4 \cite{openai2024gpt4technicalreport} and Gemini \cite{geminiteam2024geminifamilyhighlycapable} contain hundreds of billions of parameters, enabling them to learn sophisticated patterns in language and knowledge.
The training process typically involves two stages: pre-training and instruction tuning. During pre-training, models learn general language understanding and generation capabilities from vast amounts of internet text. Next, instruction tuning refines these capabilities through reinforcement learning from human feedback (RLHF), aligning the models with human preferences and task requirements \cite{ouyang2022traininglanguagemodelsfollow}.

\section{LLM Applications in Politics}
With a foundational understanding of LLMs established, we now examine their diverse applications across the political landscape. This section explores how LLMs are being employed in areas ranging from policymaking and communication to national security and legal domains, highlighting both their potential and associated challenges.

\subsection{LLMs in Legislative and Policymaking Processes}
Legislative and policymaking processes are the fundamental areas of the application. LLMs offer tools for data analysis, document classification, drafting, and participatory policy design. Their implementation, as demonstrated across various studies, highlights both their capabilities and limitations in the political arena.

In legislative contexts, LLMs have shown promise in automating the analysis and classification of policy documents. Gunes et al. \cite{gunes2023multiclassclassificationpolicydocuments} evaluated GPT-3.5 and GPT-4 on classifying U.S. Congressional bills into policy categories, achieving up to 83\% accuracy with human-computer collaboration. This hybrid approach demonstrates LLMs' potential to efficiently process large volumes of legislative texts, saving time and resources while improving accessibility for analysts.

In policymaking applications, LLMs facilitate drafting and analysis. Gao \cite{gao2023implications} highlighted LLMs’ capabilities in environmental policymaking, including real-time sentiment analysis and multilingual translation, which streamline decision-making in global governance contexts. However, these tools are prone to biases and inaccuracies, emphasizing the need for human oversight.

The use of LLMs to simulate institutional decision-making has also been explored. Zeng et al. \cite{egusphere} integrated LLMs into agent-based land-use models, simulating adaptive tax policies for regulating meat production. The study revealed that LLM-powered agents could emulate realistic institutional behaviors such as incremental policy adjustments and stakeholder negotiation. Although less efficient than optimization algorithms, these models highlight the nuanced decision-making potential of LLMs in simulating complex policy dynamics.

LLMs have been leveraged to enhance inclusivity and collaboration in policymaking. Feng et al. \cite{feng2024policyprototypingllmspluralistic} introduced a policy prototyping framework that engaged stakeholders in iterative drafting and testing of LLM-generated policies, ensuring alignment with diverse perspectives. Similarly, Kuo et al. \cite{kuo2024policycraftsupportingcollaborativeparticipatory} developed PolicyCraft, a system for collaborative policy design using case-grounded deliberation, which improved consensus and inclusivity among participants. These studies underline the importance of participatory approaches in deploying LLMs for governance.

However, challenges persist, particularly regarding bias and equity in policymaking. Ziegler et al. \cite{ziegler2024ai} examined the BBNJ Question-Answering Bot’s application in marine policy, noting that while it enhanced accessibility for under-resourced nations, it often favored perspectives from developed countries, raising concerns about fairness and representation. Addressing these biases is crucial for equitable deployment.

Finally, LLMs raise ethical concerns in lobbying activities. Nay \cite{nay2023largelanguagemodelscorporate} demonstrated their potential to streamline corporate lobbying by identifying relevant legislation and drafting persuasive communications. While these tools enhance efficiency, they also risk diminishing transparency and accountability in democratic processes, emphasizing the importance of regulatory oversight.

In summary, LLMs present powerful opportunities to augment legislative and policymaking processes, enabling efficient document processing, realistic simulations, and collaborative policy design. However, their implementation must address biases, inaccuracies, and ethical concerns to ensure responsible and equitable integration into political systems. These studies collectively underscore the potential of LLMs in politics while highlighting the critical need for thoughtful governance.

\subsection{LLMs in Political Communication and Public Opinion}
Political communication and public opinion are another aspects of the application. LLMs offer powerful new tools for analyzing and influencing political discourse. This section examines recent research on the applications of LLMs in this domain, highlighting both their potential benefits and the challenges they pose.

One of the most promising applications of LLMs is in automating the analysis of political texts. Heseltine et al. \cite{Michael20531680241236239} demonstrate the impressive accuracy of GPT-4 in annotating political texts, classifying them by political relevance, negativity, sentiment, and ideology across multiple languages. This capability could significantly reduce the need for manual coding in political science research, enabling larger-scale and more cost-effective analyses. Le Mens et al. \cite{mens2024positioningpoliticaltextslarge} further showcase the potential of LLMs to accurately position political texts in ideological and policy spaces. By querying LLMs about the positions expressed in various texts, they found high correlations with human-coded benchmarks, offering a valuable tool for mapping the ideological landscape and understanding the stances of different political actors.

In political messaging, Bai et al. \cite{voelkel2023artificial} found that AI-generated messages can be as persuasive as human-generated content in influencing attitudes on political issues, raising ethical concerns about the potential for LLMs to be used in disinformation campaigns or for targeted political advertising. Their experiments with 4,836 participants showed AI messages consistently increased policy support by 2-4 points on a 101-point scale. Hackenburg et al. \cite{Hackenburg2403116121} expanded this understanding through their large-scale experiment with 8,587 participants, revealing that LLMs' persuasive power lies primarily in their ability to craft high-quality generic messages rather than in personalization. Their finding that microtargeted messages showed no significant advantage over well-crafted generic content challenges assumptions about AI's potential for demographic-based persuasion.

However, significant concerns emerge regarding LLMs' impact on public discourse and opinion formation. Sharma et al. \cite{sharma2024generativeechochambereffects} discovered that LLM-powered conversational search systems can increase confirmatory querying by 15-16\% compared to traditional search, potentially amplifying echo chambers. When systems aligned with users' existing views, this effect intensified to 43\% confirmatory querying, suggesting LLMs might inadvertently reinforce political polarization. These results are concerning when regarded alongside Bai and Willer's demonstration of LLMs' persuasive capabilities, as they suggest AI systems could effectively persuade users while simultaneously limiting their exposure to diverse viewpoints.

Despite these challenges, research also points to ways in which the negative impacts of LLMs can be mitigated. Researchers \cite{anonymous2024linear} revealed that LLMs encode political perspectives in a "linear geometry" within their activation space, and demonstrated the ability to manipulate these representations to steer outputs towards different ideological positions. This finding opens up possibilities for monitoring and controlling bias in LLM-generated content, potentially reducing the spread of misinformation and promoting more balanced political discourse.

These studies collectively indicate that while LLMs offer powerful tools for political analysis and communication, their implementation requires careful consideration of their effects on democratic discourse. The technology's demonstrated ability to influence political attitudes while potentially amplifying echo chambers raises important questions about responsible deployment and regulation. This understanding becomes particularly crucial as political institutions increasingly consider integrating LLMs into their communication strategies and public engagement efforts.

\subsection{LLMs in Political Analysis and Collective Decision-Making}
LLMs have emerged as efficient tools in political analysis and collective decision-making, providing innovative approaches to simulate, analyze, and facilitate democratic processes. The latest research in this field underscores the potential of LLM to enhance understanding and engagement in political contexts, while also highlighting challenges that warrant further attention.

Several studies demonstrate how LLMs can simulate and predict political dynamics. Zhang et al. \cite{zhang2024electionsimmassivepopulationelection} introduced ElectionSim, a groundbreaking simulation framework leveraging LLMs to model voter behavior on a massive scale. By incorporating diverse demographic data from social media and aligning it with census datasets, ElectionSim achieved robust predictive accuracy in U.S. presidential elections, outperforming traditional agent-based models. Similarly, Sanders and Schneier \cite{sanders2023demonstrationspotentialaibasedpolitical} explored LLMs for public opinion polling, showing that GPT-3.5 could simulate ideological trends with high fidelity, offering cost-effective alternatives to traditional polling methods. Further research has examined LLMs' capacity to analyze and interpret political data. Liu et al. \cite{liu2023voicesvalidityleveraginglarge} demonstrated that GPT-4 can effectively analyze stakeholder interviews on education policy, achieving high alignment with human thematic coding and sentiment analysis. This suggests LLMs can augment human expertise in qualitative policy research, improving efficiency and consistency. 

LLMs have also been applied to decision-making scenarios. Yang et al. \cite{yang2024llmvotinghumanchoices} used GPT-4 and LLaMA-2 to simulate voting behaviors in participatory budgeting experiments. These models approximated human voting patterns, revealing biases and demonstrating potential for improving decision-making frameworks. Gudiño et al. \cite{Gudiño} extended this by proposing augmented democracy systems, where fine-tuned LLMs predicted individual and aggregate preferences, enabling citizens to unbundle policy proposals and fostering more inclusive democratic participation.

LLMs also show promise in facilitating deliberative democracy. Ashkinaze et al. \cite{ashkinaze2024pluralsguidingllmssimulated} developed Plurals, a system that leverages LLM-driven agents with personas to simulate focus groups. This approach captured diverse perspectives, fostering richer discussions compared to zero-shot generation. Tessler et al. \cite{MichaelHenryTessler} introduced the "Habermas Machine," an AI mediator that helped groups find common ground on divisive topics. The AI-mediated statements were rated higher for clarity and fairness, reducing divisions and promoting consensus.

Similarly, Huang et al. \cite{Huang3658979} proposed Collective Constitutional AI, a participatory framework that aligns LLM behavior with public principles. Models trained on public constitutions exhibited lower biases and reframed contentious topics constructively, demonstrating the potential for AI-driven inclusivity in collective decision-making.

Despite these advancements, concerns about biases and limitations remain significant. Qi et al. \cite{qi2024representationbiaspoliticalsample} highlighted representation biases in LLM simulations, with models favoring English-speaking, bipartisan democracies over multi-partisan or authoritarian regimes. Durmus et al. \cite{durmus2024measuringrepresentationsubjectiveglobal} further demonstrated that LLMs align more closely with WEIRD (Western, Educated, Industrialized, Rich, and Democratic) populations, raising questions about their applicability in global contexts.

Moreover, Herbold et al. \cite{herbold2024largelanguagemodelsimpersonate} examined the risk of impersonation, where LLMs mimicked public figures convincingly but raised misinformation concerns. Similarly, Palmer et al. \cite{palmer2023large} found that while LLMs generated persuasive political arguments, human judges distrusted AI-authored content, reflecting broader skepticism about AI's role in governance.

Despite these challenges, ongoing research highlights the potential of LLMs in political analysis and decision-making. By combining nuanced simulations, inclusive deliberative frameworks, and efforts to mitigate biases, LLMs can redefine how societies engage with complex political issues. Their role as facilitators of dialogue and tools for collective reasoning offers a promising path toward more informed and equitable democratic systems.

\subsection{LLMs in Diplomacy and National Security}
The increasing influence of LLM on Diplomacy and national security can be attributed to their capacity to enhance information processing, strategic decision-making, and operational efficiency. Their capacity to analyze massive data, simulate complex scenarios, and facilitate communication positions them as valuable tools in these high-stakes domains.

One significant application of LLMs is in enhancing military planning and intelligence analysis. For instance, Defense Llama, developed by Scale AI in collaboration with Meta, is a specialized LLM tailored for U.S. national security applications \cite{defense_llama}. By fine-tuning Meta’s Llama 3 for defense use cases, Defense Llama integrates into command platforms and decision-support systems, aiding defense operations securely. This demonstrates how LLMs can be adapted to meet specific national security needs, improving decision-making processes and operational readiness.

LLMs have also been utilized to simulate historical battles and international conflicts, providing deeper insights into strategic decision-making and conflict dynamics. BattleAgent, introduced by Lin et al. \cite{lin2024battleagentmultimodaldynamicemulation}, merges large vision-language models with multi-agent systems to dynamically emulate historical battles. By modeling complex interactions among agents and their environments, BattleAgent captures both strategic decisions and individual participant experiences, offering a granular understanding of historical events that traditional analyses might overlook. Extending this approach to broader international conflicts, WarAgent, developed by Hua et al. \cite{hua2023war}, leverages LLMs to simulate historical wars such as World Wars I and II. By modeling the decision-making processes of nations, WarAgent explores triggers and conditions leading to wars, highlighting the potential of AI to provide data-driven insights into conflict resolution and international relations. These simulations not only enhance our understanding of past conflicts but also offer valuable tools for policymakers to anticipate and mitigate future crises.

In the realm of Diplomacy, generative AI technologies are reshaping global diplomatic practices. Bano et al. \cite{bano2023rolegenerativeaiglobal} evaluate the impact of generative AI on Diplomacy and propose a strategic framework for integrating these technologies into modern diplomatic practices. Generative AI enhances public Diplomacy by enabling nuanced audience engagement and personalized messaging. It also improves negotiation and crisis management through tools like sentiment analysis and real-time analytics. This integration fosters more effective communication and collaboration among nations, potentially leading to more peaceful and cooperative international relations.

LLMs also show promise in facilitating cooperative behaviors in complex multi-agent environments. Mukobi et al. \cite{mukobi2023welfarediplomacybenchmarkinglanguage} introduce Welfare Diplomacy, a modified version of the board game Diplomacy designed to evaluate and enhance cooperative capabilities in AI systems. Using language model agents such as GPT-4, the study benchmarks cooperative behaviors and explores vulnerabilities to exploitation. The findings suggest that while LLMs can achieve high social welfare by demilitarizing and cooperating, they remain susceptible to defections, highlighting the importance of developing robust strategies for maintaining cooperation in adversarial settings.

Despite these positive applications, the deployment of LLMs in Diplomacy and national security also presents significant risks and challenges. Rivera et al. \cite{rivera2024escalation} investigate the behavior of LLMs in simulated wargames, revealing tendencies for escalation, unpredictable spikes in aggressive actions, and arms-race dynamics. Notably, models occasionally chose violent or nuclear escalatory actions even in neutral scenarios. These findings highlight the risks of employing LLMs in military settings without robust safeguards and emphasize the necessity for comprehensive evaluation and ethical considerations before deployment.

Moreover, Caballero et al. \cite{caballero2024largelanguagemodelsnational} discuss the limitations of LLMs in high-stakes contexts due to issues like hallucinations, data privacy concerns, and vulnerability to adversarial attacks. While LLMs can automate tasks and enhance data analysis, their reliability in critical decision-making processes is not yet assured. The potential for AI deception, as explored by Park et al. \cite{park2023aideceptionsurveyexamples}, further underscores the risks associated with AI systems developing deceptive capabilities, including the inducement of false beliefs, which can lead to fraud, loss of control, and destabilization in international relations.

To address these challenges, researchers emphasize the importance of developing safeguards, ethical guidelines, and robust frameworks for integrating LLMs into national security applications. Coupling LLMs with decision-theoretic principles, as suggested by Caballero et al. \cite{caballero2024largelanguagemodelsnational}, can facilitate actionable decisions while mitigating risks. Additionally, ongoing research into AI deception detection and mitigation strategies, as proposed by Park et al. \cite{park2023aideceptionsurveyexamples}, is crucial for ensuring the responsible deployment of AI technologies.

Overall, the integration of LLMs into Diplomacy and national security holds significant potential for enhancing decision-making, fostering cooperation, and improving operational efficiency. By carefully addressing the associated risks and challenges, these technologies can contribute to more effective and secure global interactions.

\subsection{LLMs as Simulated Agents in Economic and Social Models}
LLMs are employed to simulate intricate social and economic systems, thereby suggesting novel avenues for comprehending human behavior and policy impacts. This section examines recent research on LLMs in economic and social modeling, emphasizing their capacity to reshape research methodologies and inform decision-making.

Several studies have demonstrated the ability of LLMs to simulate human-like agents in social and economic scenarios. Park et al. \cite{park2022socialsimulacracreatingpopulated} introduced "social simulacra," a technique using LLMs to generate plausible user interactions in social computing systems. Their SimReddit platform allows designers to explore and refine community designs by simulating large-scale user behavior. Similarly, Wang et al. \cite{wang-etal-2023-humanoid} developed "Humanoid Agents," integrating System 1 thinking processes like emotions and relationships into LLMs to create more realistic and adaptive agents in simulated environments. These studies showcase the potential of LLMs to capture nuanced social dynamics.

Building on this foundation, researchers have explored the use of LLMs in simulating specific social and economic phenomena. Piatti et al. \cite{piatti2024cooperatecollapseemergencesustainable} investigated sustainable cooperation in multi-agent systems using LLMs. Their GOVSIM platform simulates resource-sharing dilemmas, revealing that advanced LLMs, particularly when combined with communication and moral reasoning capabilities, can achieve partial sustainability in managing shared resources. In the realm of public health, Williams et al. \cite{williams2023epidemicmodelinggenerativeagents} introduced a generative agent-based modeling (GABM) approach to simulate epidemic spread. Their research demonstrated that LLMs can effectively model human behaviors like self-isolation and quarantine, leading to more realistic epidemic patterns and improved policy planning capabilities.

The application of LLMs extends to simulating human decision-making in economic contexts. Horton \cite{horton2023largelanguagemodelssimulated} positioned LLMs as "Homo Silicus," computational analogs of humans for economic experiments. By replicating classic behavioral economics studies, the research showed that LLMs can exhibit context-sensitive behavior qualitatively similar to human data, offering a cost-effective and ethical alternative for piloting experiments and exploring economic theories. Kaashoek and Horton \cite{Kaashoek2024Impact} further explored the implications of LLMs on labor market matching, emphasizing their potential to improve information processing and decision-making for both job seekers and employers. However, they also cautioned against potential risks like signal homogenization and AI biases.

Beyond individual decision-making, LLMs are also being used to model social systems at scale. Xiao et al. \cite{xiao2023simulatingpublicadministrationcrisis} introduced a GABSS framework to simulate public administration crises, demonstrating how LLM-driven agents can adapt and respond to events like water pollution incidents based on memory and interactions. Ghaffarzadegan et al. \cite{Ghaffarzadegan_2024} presented a GABM approach to simulate norm diffusion, highlighting the emergent dynamics and sensitivity to initial conditions in LLM-driven simulations. Jiang et al. \cite{jiang2023socialllmmodelinguserbehavior} developed Social-LLM, a scalable framework integrating LLM embeddings with social network data to predict user behaviors like political polarization and hate speech.

The collective findings of these studies illustrate the adaptability of LLMs in economic and social modeling. By emulating human-like reasoning, social interactions, and emergent behaviors, LLMs present a formidable instrument for comprehending intricate systems, forecasting outcomes, and influencing policy decisions. Nevertheless, it is imperative to confront constraints such as bias, prompt sensitivity, and interpretability challenges through continuous investigation and ethical development.

\subsection{LLMs in Legal Applications}
The potential of LLMs to affect significant changes in various areas of the legal domain is being investigated. These areas include legal research and drafting, judicial decision-making, and access to justice initiatives. This section examines recent research on LLMs in legal applications, highlighting their capabilities, limitations, and potential implications for the legal profession and society at large.

Several studies have investigated the performance of LLMs on standardized legal tasks, such as the bar exam. Katz et al. \cite{katz2024gpt} demonstrated that GPT-4 achieved a passing score on the Uniform Bar Examination (UBE), surpassing prior models and human averages in multiple-choice and essay components. This suggests that LLMs are capable of applying complex legal principles and reasoning, although challenges remain in nuanced tasks like statutory interpretation and policy analysis.

Specialized benchmarks have been developed to assess LLMs' legal reasoning abilities more comprehensively. Guha et al. \cite{guha2024legalbench} introduced LegalBench, a collaborative benchmark encompassing 162 tasks across six categories of legal reasoning. Their evaluation of 20 LLMs revealed significant performance variability, with GPT-4 excelling in rule-application and rule-conclusion tasks. Similarly, Fei et al. \cite{fei2023lawbench} developed LawBench, a benchmark focused on the Chinese civil law system, highlighting the challenges in developing robust legal AI systems for diverse legal traditions.

While LLMs show promise in legal applications, their limitations must be carefully considered. Dahl et al. \cite{dahl2024large} conducted a systematic analysis of "legal hallucinations" in LLMs, where generated outputs deviate from legal facts. Their findings revealed high hallucination rates across various LLMs, influenced by factors such as jurisdiction, court level, and case salience. These hallucinations raise concerns over the reliability of LLMs in unsupervised legal tasks and emphasize the need for human oversight.

Despite these challenges, ongoing research is exploring strategies to enhance the reliability of LLMs in legal contexts. Colombo et al. \cite{colombo2024saullm} introduced SaulLM-54B and SaulLM-141B, two large language models specifically adapted for the legal domain. Their study demonstrated the benefits of scaling and domain-specific tuning, achieving state-of-the-art performance on LegalBench-Instruct. Harrington \cite{harrington2023case} explored the potential of law-specific LLMs like Casetext's Cocounsel and Westlaw AI, highlighting the advantages of curated legal databases and advanced tools like embeddings and multi-stage prompting.

The integration of LLMs into legal workflows raises important questions about the future of the legal profession and access to justice. Cheong et al. \cite{cheong2024not} engaged legal experts in workshops to examine the ethical implications of LLMs providing legal advice. Their findings emphasize the need for nuanced guidelines and responsible deployment strategies, focusing on guiding users to relevant information rather than offering definitive legal opinions. Shui et al. \cite{shui2023comprehensive} evaluated LLMs on legal judgment prediction, demonstrating the value of integrating similar case demonstrations and label candidates to prompt LLMs effectively.

These studies demonstrate LLMs hold significant potential to transform legal applications, but their limitations and ethical implications must be addressed through ongoing research and collaboration between the legal and AI communities. While LLMs may not replace human lawyers entirely, they can serve as tools to improve legal research, drafting, and decision-making, potentially improving efficiency and access to justice. However, ensuring responsible development and deployment of LLMs in legal contexts is crucial to mitigate risks and maximize their benefits for society.

\section{Future Prospects}
The evolution of LLMs in political applications presents both possibilities and challenges for the future of governance, democracy, and international relations. This section explores key trends and potential developments across several critical domains.

\textbf{AI-Facilitated Deliberation:} As demonstrated by Tessler et al. \cite{MichaelHenryTessler}, AI mediators may evolve to facilitate more effective group discussions on contentious issues. These systems could help bridge ideological divides and foster consensus-building in increasingly polarized societies.

\textbf{Human-AI Collaboration:} Fostering collaborative frameworks that leverage both human expertise and LLM capabilities like PolicyCraft \cite{kuo2024policycraftsupportingcollaborativeparticipatory} and Policy Prototyping \cite{feng2024policyprototypingllmspluralistic} will be essential for navigating complex political challenges and ensuring human oversight in critical decision-making.

\textbf{Dynamic Policy Simulation:} Future systems may enable more sophisticated modeling of policy impacts, expanding on current frameworks for simulating institutional decision-making \cite{egusphere, naveed2023comprehensive, Ghaffarzadegan_2024}.

\textbf{Bias Mitigation:} Ongoing research should prioritize developing robust methods for identifying and mitigating biases in LLM training data and outputs to ensure fairness and equity in political 
applications \cite{qi2024representationbiaspoliticalsample}.

\textbf{Transparency and Accountability:} Establishing clear guidelines for transparency in LLM development and deployment, along with mechanisms for accountability in cases of misinformation or harmful outputs, is crucial for responsible use.

\section{Conclusion}
In this survey, we provide a comprehensive overview of LLMs in politics, from policymaking and public opinion analysis to international relations and legal domains. LLMs offer unprecedented capabilities in automating tasks, analyzing large datasets, and facilitating human-like communication, with the potential to enhance efficiency, inclusivity, and understanding in political processes. However, challenges persist, particularly regarding bias mitigation, transparency, and accountability. 

Future research should prioritize developing robust methods for evaluating and mitigating biases in LLMs, ensuring fairness and equity in political applications.  

Fostering interdisciplinary collaborations between AI researchers, social scientists, and policymakers is crucial for addressing these challenges and guiding the responsible development and deployment of LLMs in political contexts.  

The evolving landscape of LLMs in politics necessitates ongoing critical engagement to harness their transformative potential while safeguarding democratic values and promoting a more just and equitable society.

{\small
\bibliographystyle{plain}
\bibliography{egbib}
}

\end{document}